\pdfoutput=1

\documentclass[11pt]{article}

\usepackage{ACL2023}

\usepackage{times}
\usepackage{latexsym}

\usepackage[T1]{fontenc}

\usepackage[utf8]{inputenc}

\usepackage{microtype}

\usepackage{inconsolata}

\usepackage{graphicx}
\usepackage{lipsum}
\usepackage{svg}
\usepackage{booktabs}
\usepackage{highlight}
\usepackage{amsmath}

\title{WangLab at MEDIQA-Chat 2023: Clinical Note Generation from Doctor-Patient Conversations using Large Language Models}

\author{
John Giorgi$^{1,2,3\ast}$ \quad Augustin Toma$^{1,3,4}$\thanks{\enspace Core contributors. See \hyperref[sec:contrib]{author contributions}} \quad Ronald Xie$^{1,2,3,4\ast}$ \\ \textbf{Sondra~S.~Chen}$^{1,5}$ \quad \textbf{Kevin~R.~An}$^{1,6}$ \quad \textbf{Grace~X.~Zheng}$^{1,5}$ \quad \textbf{Bo Wang}$^{1,3,4\ast}$\vspace{5pt} \\
  $^{1}$University of Toronto \quad
  $^{2}$Terrence Donnelly Centre for Cellular and Biomolecular Research \\
  $^{3}$Vector Institute for AI \quad
  $^{4}$University Health Network \quad
  $^{5}$Sunnybrook Health Sciences Centre\\
  $^{6}$Department of Cardiac Surgery, University of Toronto\vspace{5pt}\\
  {\footnotesize \bf\texttt{\{john.giorgi, augustin.toma, ronald.xie, bowang.wang\}@mail.utoronto.ca}}
}

\begin{document}
\maketitle

\begin{abstract}
This paper describes our submission to the MEDIQA-Chat 2023 shared task for automatic clinical note generation from doctor-patient conversations. We report results for two approaches: the first fine-tunes a pre-trained language model (PLM) on the shared task data, and the second uses few-shot in-context learning (ICL) with a large language model (LLM). Both achieve high performance as measured by automatic metrics (e.g. ROUGE, BERTScore) and ranked second and first, respectively, of all submissions to the shared task. Expert human scrutiny indicates that notes generated via the ICL-based approach with GPT-4 are preferred about as often as human-written notes, making it a promising path toward automated note generation from doctor-patient conversations.\footnote{\url{https://github.com/bowang-lab/MEDIQA-Chat-2023}}
\end{abstract}

\section{Introduction}

\begin{figure}[t]
\includegraphics[width=\columnwidth]{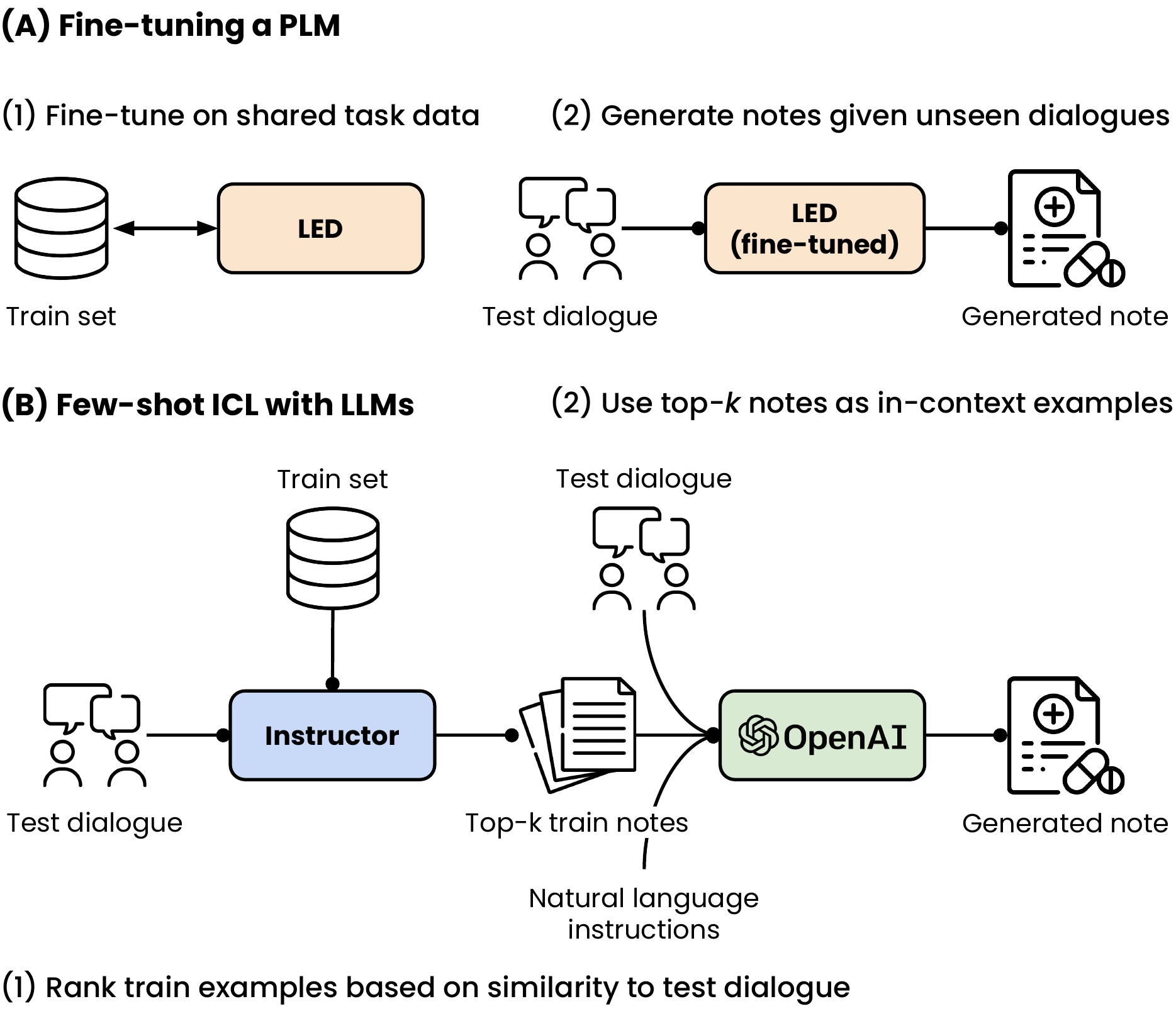}
\caption{(\textbf{A}) Fine-tuning a pre-trained language model (PLM), Longformer-Encoder-Decoder (LED, \citealt{Beltagy2020LongformerTL}). (\textbf{B}) In-context learning (ICL) with large language models (LLMs). We rank train examples based on their similarity to the test dialogue using Instructor \citep{Su2022OneEA}. Notes of the top-\(k\) most similar examples are then used as in-context examples to form a prompt alongside natural language instructions and fed to GPT-4 \citep{OpenAI2023GPT4TR} to generate the clinical note.}
\label{fig:overview}
\vspace{-2.7mm}
\end{figure}

The growing burden of clinical documentation has emerged as a critical issue in healthcare, increasing job dissatisfaction and burnout rates among clinicians and negatively impacting patient experiences \citep{Friedberg2013FactorsAP, Babbott2014ElectronicMR, Arndt2017TetheredTT}. On the other hand, timely and accurate documentation of patient encounters is critical for safe, effective care and communication between specialists. Therefore, interest in assisting clinicians by automatically generating consultation notes is mounting \citep{finley-etal-2018-automated, enarvi-etal-2020-generating, Molenaar2020MedicalDS, knoll-etal-2022-user}.

To further encourage research on automatic clinical note generation from doctor-patient conversations, the MEDIQA-Chat Dialogue2Note shared task was proposed \citep{mediqa-chat-2023}. Here, we describe our submission to subtask B: the generation of full clinical notes from doctor-patient dialogues. We explored two approaches; the first fine-tunes a pre-trained language model (PLM, \textsection \ref{fine-tuning}), while the second uses few-shot in-context learning (ICL, \textsection \ref{in-context-learning}). Both achieve high performance as measured by automatic natural language generation metrics (\textsection \ref{results}) and ranked second and first, respectively, of all submissions to the shared task. In a human evaluation with three expert physicians, notes generated via the ICL-based approach with GPT-4 were preferred about as often as human-written notes (\textsection \ref{results:human-eval}).

\begin{figure*}[t]
\centering
\includegraphics[width=\textwidth]{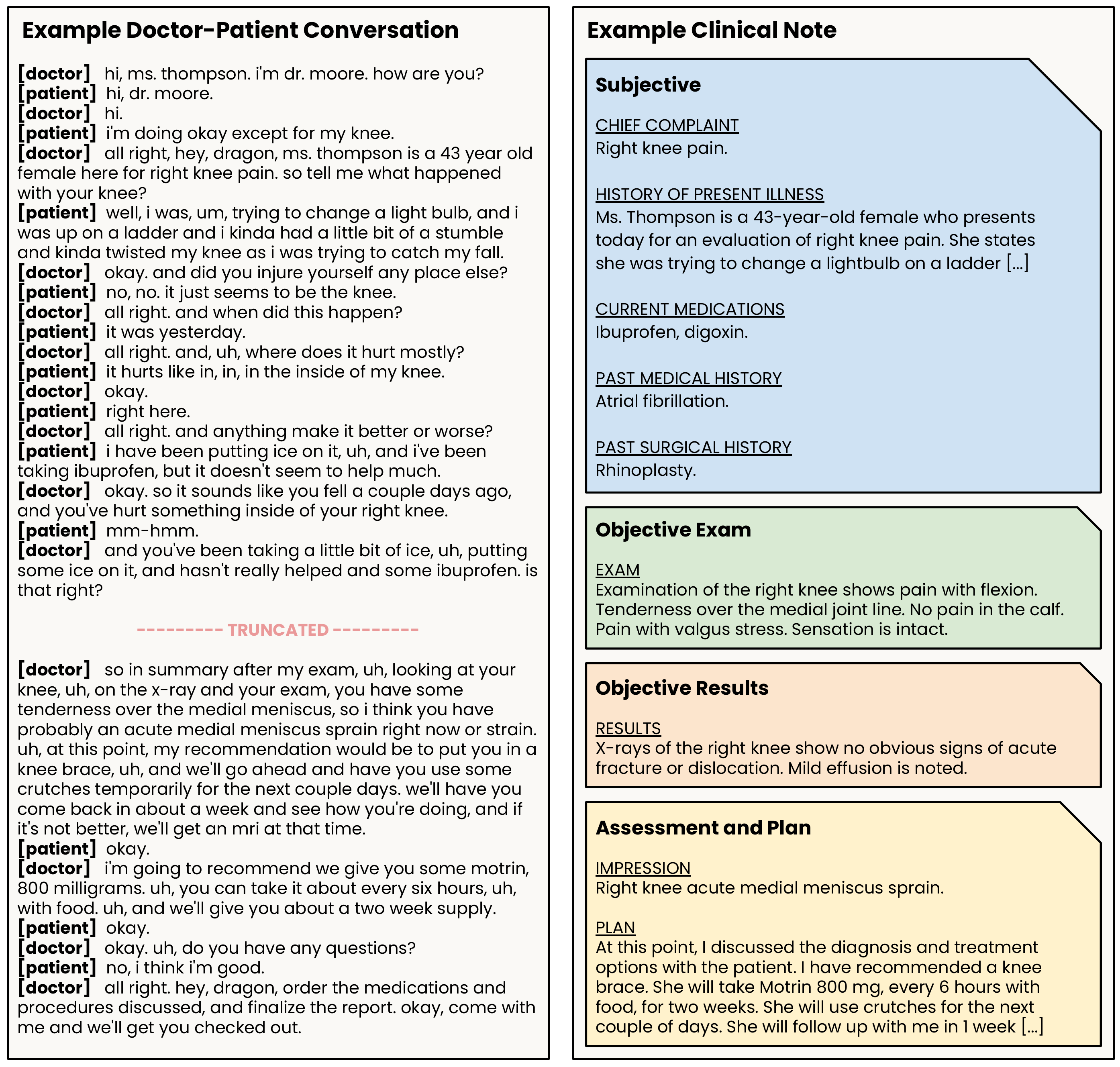}
\caption{Example of a paired doctor-patient conversation and clinical note from the subtask B validation set. Dialogue has been lightly cleaned for legibility (e.g. remove trailing white space). Parts of the dialogue and note have been truncated. During evaluation, sections are grouped under one of four categories: ``Subjective'', ``Objective Exam'', ``Objective Results'', and ``Assessment and Plan'' (see \textsection \ref{shared-task} for details).}
\label{fig:full-example}
\vspace{-2.9mm}
\end{figure*} 

\section{Shared Task and Dataset}

MEDIQA-Chat 2023 proposed two shared tasks:

\begin{enumerate}
\item{\textbf{Dialogue2Note Summarization}: Given a conversation between a doctor and patient, the task is to produce a clinical note summarizing the conversation with one or more note sections (e.g. Assessment, Past Medical History).}
\item{\textbf{Note2Dialogue Generation}: Given a clinical note, the task is to generate a synthetic doctor-patient conversation related to the information described in the note.}
\end{enumerate}

\noindent We focused on Dialogue2Note, which is divided into two subtasks. In subtask `A' \citep{ben-abacha-etal-2023-empirical}, the goal is to generate \textit{specific sections} of a note given partial doctor-patient dialogues. In subtask `B' \citep{aci-demo}, the goal is \textit{full} note generation from complete dialogues. The remainder of the paper focuses on subtask B; see \autoref{appendix:subtask-a} for our approach to subtask A, which also ranks first of all submissions to the shared task. 

\subsection{Task definition}
\label{shared-task}

Each of the \(k\) examples consist of a doctor-patient dialogue, \(D = {d_1, \dots, d_k}\) and a corresponding clinical note, \(N = {n_1, \dots, n_k}\). The aim is to automatically generate a note \(n_i\) given a dialogue \(d_i\). Each note comprises one or more sections, such as ``Chief Complaint'', and ``Family history''. During evaluation, sections are grouped under one of four categories: ``Subjective'', ``Objective Exam'', ``Objective Results'', and ``Assessment and Plan''.\footnote{See \href{https://github.com/abachaa/MEDIQA-Chat-2023/blob/main/scripts/sectiontagger.py}{here} for the mapping} See \autoref{fig:full-example} for an example doctor-patient conversation and clinical note pair.

\subsection{Dataset}
\label{dataset}

The dataset comprises 67 train and 20 validation examples, featuring transcribed dialogues from doctor-patient encounters and the resulting clinician-written notes. Each example is labelled with the `dataset source', indicating the dialogue transcription system used to produce the note. 

\section{Approach}

We take two high-performant approaches to the shared task. In the first, we fine-tune a pre-trained language model (PLM) on the provided training set (\textsection \ref{fine-tuning}). In the second, we use in-context learning (ICL) with a large language model (LLM, \textsection \ref{in-context-learning}).

\begin{figure}[t]
\centering
\includegraphics[width=\columnwidth]{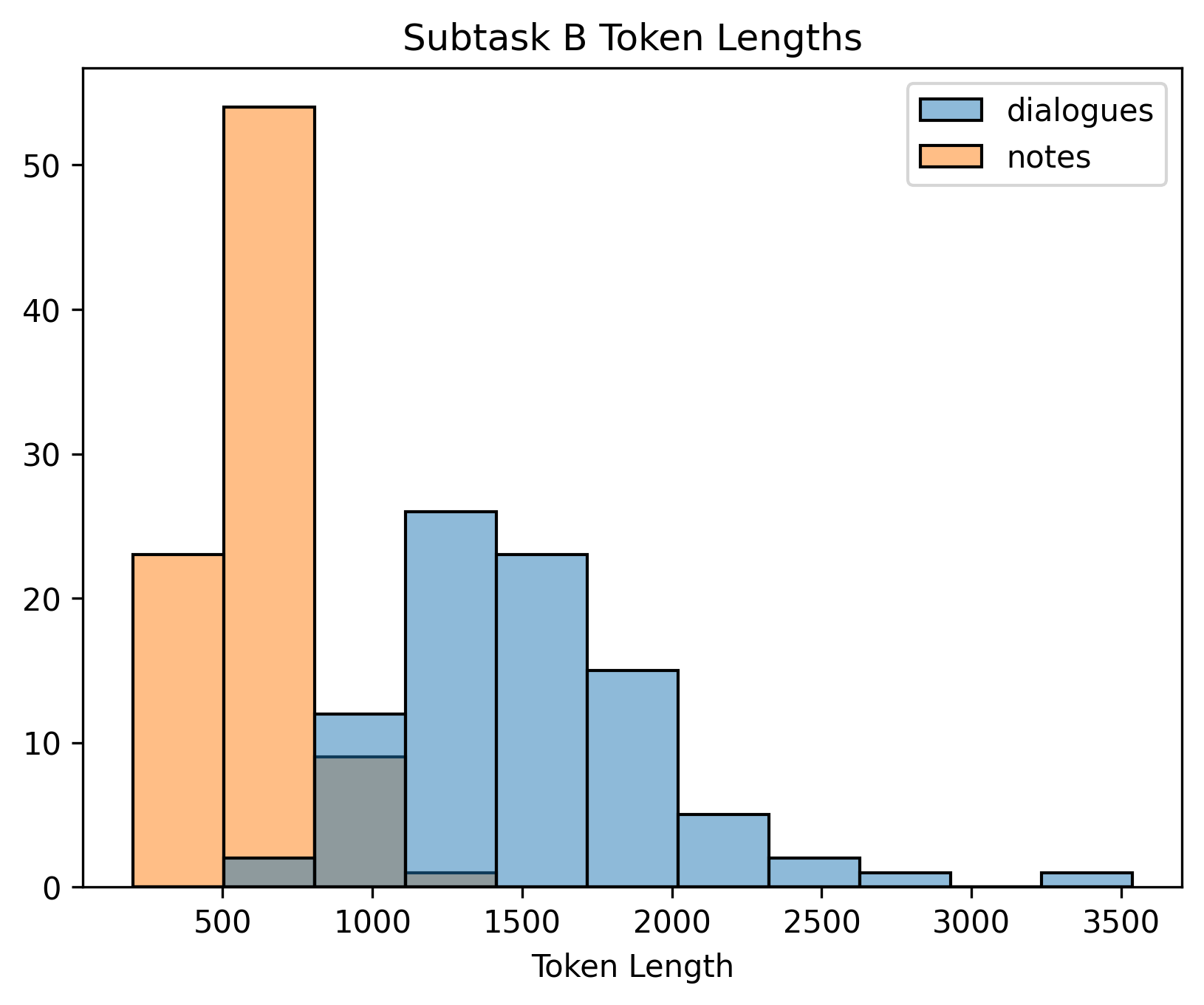}
\caption{Histogram of token lengths for subtask B train and validation sets. Dialogues and notes were tokenized with \texttt{\href{http://github.com/openai/tiktoken}{tiktoken}} using the ``\texttt{gpt-4}'' encoding.}
\label{fig:taskb-lengths}
\end{figure}

\begin{figure}[t]
\includegraphics[width=\columnwidth]{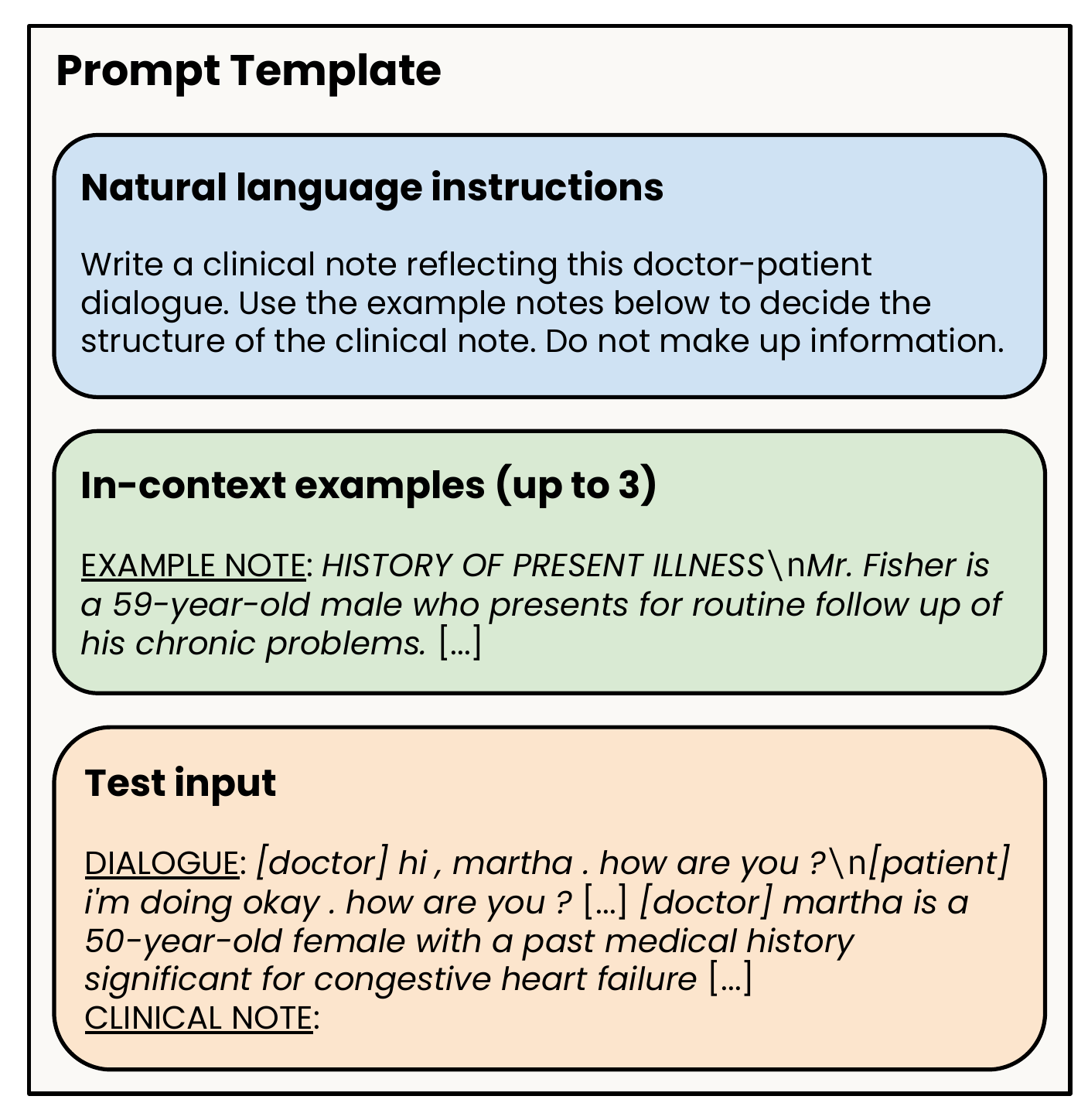}
\caption{Prompt template for our in-context learning (ICL) based approach. Each prompt includes natural language instructions, up to 3 in-context examples, and an unseen doctor-patient dialogue as input.}
\label{fig:prompt}
\end{figure}

\subsection{Fine-tuning pre-trained language models}
\label{fine-tuning}

As a first approach, we fine-tune a PLM on the training set following a canonical, sequence-to-sequence training process (\autoref{fig:overview}  A; see \autoref{seq2seq} for details). Given the length of input dialogues (\autoref{fig:taskb-lengths}), we elected to use Longformer-Encoder-Decoder (LED, \citealt{Beltagy2020LongformerTL}), which has a maximum input size of 16,384 tokens. We begin fine-tuning from a LED\textsubscript{LARGE} checkpoint tuned on the PubMed summarization dataset \citep{cohan-etal-2018-discourse}, which performed best in preliminary experiments.\footnote{\url{https://huggingface.co/patrickvonplaten/led-large-16384-pubmed}} The model was fine-tuned using HuggingFace Transformers \citep{wolf-etal-2020-transformers} on a single NVIDIA A100-40GB GPU. Hyperparameters were lightly tuned on the validation set.\footnote{See Appendix \ref{appendix:led-hyperparams} for details}

\subsection{In-context learning with LLMs} 
\label{in-context-learning}

As a second approach, we attempt subtask B with ICL. We chose GPT-4 \citep{OpenAI2023GPT4TR}\footnote{Specifically, the 03/14/2023 snapshot, ``\href{https://platform.openai.com/docs/models/gpt-4}{\texttt{gpt-4-0314}}''} as the LLM and designed a simple prompt, which included natural language instructions and in-context examples (\autoref{fig:prompt}). We limited the prompt size to 6192 tokens --- allowing for 2000 output tokens, as the model's maximum token size is 8,192 --- and used as many in-context examples as would fit within this token limit, up to a maximum of 3. We set the \texttt{temperature} parameter to 0.2 and left all other \href{https://platform.openai.com/docs/api-reference/completions}{hyperparmeters of the OpenAI API} at their defaults.

\paragraph{Natural language instructions}

During preliminary experiments, we found that GPT-4 was not overly sensitive to the exact phrasing of the natural language instructions in the prompt. We, therefore, elected to use short, simple instructions (\autoref{fig:prompt}).

\paragraph{In-context example selection}

Each in-context example is a note from the train set. To select the notes, we first embed the dialogues of each training example and the input dialogue. Train dialogues are then ranked based on cosine similarity to the input dialogue; notes of the resulting top-\(k\) training examples are selected as the in-context examples (see \autoref{fig:overview}, B). Dialogues were embedded using Instructor \citep{Su2022OneEA}, a text encoder that supports natural language instructions.\footnote{We used the following instructions: ``Represent the Medicine dialogue for clustering: \texttt{\{dialogue\}}''} Lastly, we restricted in-context examples to be of the same `dataset source' (see \textsection \ref{dataset}) as the input dialogue, hypothesizing that this may improve performance.\footnote{Manual review revealed that dataset source was predictive of note structure \& style; likely because it indicates which clinician or electronic health record system produced the note}

\subsection{Evaluation}
\label{evaluation}

Models are evaluated with the official evaluation script\footnote{\url{https://github.com/abachaa/MEDIQA-Chat-2023/blob/main/scripts/evaluate_summarization.py}} on the validation set (as test notes are not provided). Generated notes are evaluated against the provided ground truth notes with ROUGE \citep{lin-2004-rouge}, BERTScore \citep{bertscore} and BLEURT \citep{sellam-etal-2020-bleurt}. We report performance as the arithmetic mean of ROUGE-1 F1, BERTScore F1 and BLEURT-20 \citep{pu-etal-2021-learning}. 

\section{Results}
\label{results}

\subsection{Fine-tuning pre-trained language models}
\label{results:fine-tuning}

We present the results of fine-tuning LED in \autoref{tab:led-results}. Due to the non-determinism of the LED implementation,\footnote{\url{https://github.com/huggingface/transformers/issues/12482}} we report the mean results of three training runs. Unsurprisingly, we find that scaling the model size from LED\textsubscript{BASE} (12 layers, \(\sim\)162M parameters) to LED\textsubscript{LARGE} (24 layers, \(\sim\)460M parameters) leads to sizable gains in performance. Performance further improves by initializing the model with a checkpoint fine-tuned on the PubMed summarization dataset (LED\textsubscript{LARGE-PubMed}). This is likely because (1) Dialouge2Note resembles a summarization task, and (2) text from PubMed is more similar to clinical text than is the general domain text used to pre-train LED.\footnote{LED is initialized from BART \citep{lewis-etal-2020-bart}, which was pre-trained on a combination of text from Wikipedia and BooksCorpus \citep{Zhu_2015_ICCV}} Our submission to the shared task using this approach ranked second overall, outperforming the next-best submission by 2.7 average score; a difference comparable to the improvement in performance we see by doubling model size (see LED\textsubscript{BASE} vs. LED\textsubscript{LARGE}, \autoref{tab:led-results}).

\begin{table}[t]
\small
\centering
\caption{Fine-tuning LED. Mean and standard deviation (SD) of three training runs is shown. Scaling model size and pre-training on a related task improve performance. \textbf{Bold}: best scores.}
\label{tab:led-results}
\resizebox{\columnwidth}{!}{%
\begin{tabular}{@{}lcccc@{}}
\toprule
Model                    & ROUGE-1 F1                & BERTScore F1              & BLEURT                  & Avg. \\ \midrule
LED\textsubscript{BASE}  & 57.0\textsubscript{0.4} & 67.3\textsubscript{0.1} & 36.9\textsubscript{0.0} & 53.8 \\
LED\textsubscript{LARGE} & 59.8\textsubscript{0.2} & 70.0\textsubscript{0.6} & 41.1\textsubscript{0.8} & 57.0 \\
LED\textsubscript{LARGE-PubMed} & \textbf{61.7\textsubscript{0.4}} & \textbf{70.7\textsubscript{0.2}} & \textbf{41.5\textsubscript{0.6}} & \textbf{57.9} \\ \bottomrule
\end{tabular}%
}
\end{table}

\begin{table*}[t]
\small
\centering
\caption{ICL with GPT-4. Mean of ROUGE-1 F1, BERTScore F1 and BLEURT for three runs is shown. Selecting in-context examples based on similarity to input dialogue improves performance. Dialogue-note pairs as in-context examples (omitting 3-shot results due to token length limits) underperforms notes only. Filtering in-context examples to be of the same `dataset source' as the input dialogue has little effect. \textbf{Bold}: best scores. SD < 0.1 in all cases.}
\label{tab:icl-results}
\begin{tabular}{@{}ccccclcccc@{}}
\toprule
                             & \multicolumn{4}{c}{Unfiltered}                               &  & \multicolumn{4}{c}{Filtered by dataset source}      \\ \cmidrule(lr){2-5} \cmidrule(l){7-10} 
Example selection strategy & 0-shot          & 1-shot          & 2-shot          & 3-shot &  & 0-shot & 1-shot          & 2-shot          & 3-shot \\ \midrule
\multicolumn{10}{c}{\textit{Dialogue-note pairs as in-context examples}}                                                                             \\ \midrule
random                       & \gradient{52.2} & \gradient{54.5} & \gradient{53.9} & --     &  & --     & \gradient{54.8} & \gradient{54.5} & --     \\
similar dialogues            & --              & \gradient{59.4} & \gradient{59.4} & --     &  & --     & \gradient{60.1} & \gradient{60.3} & --     \\ \midrule
\multicolumn{10}{c}{\textit{Notes only as in-context examples}}                                                                                      \\ \midrule
random            & -- & \gradient{56.3} & \gradient{56.7} & \gradient{56.7} &  & -- & \gradient{56.3}          & \gradient{56.5} & \gradient{56.7}          \\
similar dialogues & -- & \gradient{60.7} & \gradient{60.6} & \gradient{60.4} &  & -- & \gradientbold{60.8} & \gradient{60.4} & \gradientbold{60.8} \\ \bottomrule
\end{tabular}
\end{table*}

\begin{table}[t]
\centering
\caption{Human evaluation. Three physicians selected their preference from human written ground-truth notes (GT), notes produced by the fine-tuned model (FT) and notes produced by in-context learning (ICL). Win rate is \% of cases where note was preferred, excluding ties.}
\label{tab:human-eval}
\resizebox{\columnwidth}{!}{%
\begin{tabular}{@{}ccccccccccc@{}}
\toprule
               & \multicolumn{3}{c}{Preferred} &  & \multicolumn{2}{c}{Ties} &  & \multicolumn{3}{c}{Win rate (\%)} \\ \cmidrule(lr){2-4} \cmidrule(lr){6-7} \cmidrule(l){9-11} 
Physician      & GT       & FT      & ICL      &  & FT/ICL        & All      &  & GT      & FT      & ICL      \\ \midrule
1              & 9        & 1       & 4        &  & 2             & 4        &  & \gradientwinrate{64}      & \gradientwinrate{7}    & \gradientwinrate{29}       \\
2              & 5        & 0       & 14       &  & 0             & 1        &  & \gradientwinrate{26}      & \gradientwinrate{0}       & \gradientwinrate{74}       \\
3              & 9        & 0       & 6        &  & 0             & 5        &  & \gradientwinrate{60}      & \gradientwinrate{0}       & \gradientwinrate{40}       \\ \midrule
\textbf{Total} & 23       & 1       & 24       &  & 2             & 10        &  & \gradientwinrate{48}      & \gradientwinrate{2}       & \gradientwinrate{50}       \\ \bottomrule
\end{tabular}%
}
\vspace{-3mm}
\end{table}

\subsection{In-context learning with LLMs} 
\label{results:in-context-learning}

We present the results of ICL with GPT-4 in \autoref{tab:icl-results}. We note several interesting trends in order of magnitude of impact. First, \textit{selecting in-context examples based on the similarity of dialogues has a strong positive impact}, typically improving average score by 4 or more. \textit{Using only notes as in-context examples, as opposed to dialogue-note pairs, also has a positive impact}, typically improving average score by \(\sim\)1. Surprisingly, \textit{increasing the number of in-context examples had a marginal effect on performance}. Together these results suggest that the in-context examples' primary benefit is providing guidance with regard to the expected note structure, style and length. Finally, \textit{filtering in-context examples to be of the same `dataset source' as the input dialogue has a negligible impact on performance}.

The best strategy out-performs LED by almost 3 average score (60.8 vs. 57.9, see \autoref{tab:led-results} \& \autoref{tab:icl-results}) and achieves first place of all submissions to the shared task, out-performing the runner up by \(>9\) average score. We conclude that (1) few-shot ICL with GPT-4, using as little as one example, is a performant approach for note generation from doctor-patient conversations, and (2) using the notes of semantically similar dialogue-note pairs is a strong strategy for selecting the in-context examples.

\subsection{Human evaluation} 
\label{results:human-eval}

Automatic evaluation metrics like ROUGE, BERTScore and BLEURT are imperfect and may not correlate with aspects of human judgment.\footnote{See \textsection \ref{limitations} for an extended discussion} Therefore, we conducted an expert human evaluation to validate our results. To make annotation feasible, we conducted it on the validation set (20 examples) using the best performing fine-tuned model: LED\textsubscript{LARGE-PubMed} (\autoref{tab:led-results}), and best performing ICL-based approach: 3-shot, similar, note-only examples filtered by dataset type (\autoref{tab:icl-results}).

Three senior resident physicians\footnote{The three annotators are a subset of the authors who did not interact with the model or model outputs before annotation} were shown a ground truth note, a note generated by the fine-tuned model, and a note generated by the ICL-based approach for each example (presented in random order as clinical note `A', `B' and `C') and asked to select which note(s) they preferred, given a dialogue and some simple instructions:\\

\noindent \textbf{Instructions}: Please asses the \textbf{clinical notes A}, \textbf{B} and \textbf{C} relative to the \textbf{provided doctor-patient dialogue}. For each set of notes, you should select which note you prefer (`A', `B', or `C'). If you have approximately equal preference for two notes, select (`A/B', `B/C', or `C/A'). If you have no preference, select `A/B/C'. A `good' note should contain \textit{all} \textbf{critical}, \textit{most} \textbf{non-critical} and \textit{very little} \textbf{irrelevant} information mentioned in a dialogue:

\begin{itemize}
    \item \textbf{Critical}: Items medico-legally required to document the diagnosis and treatment decisions whose absence or incorrectness may lead to wrong diagnosis and treatment later on, e.g. the symptom "cough" in a suspected chest infection consultation. This is the key information a note needs to capture correctly in order to not mislead clinicians.
    \item \textbf{Non-critical}: Items that should be documented in a complete note but whose absence will not affect future treatment or diagnosis, e.g. "who the patient lives with" in a consultation about chest infection.
    \item \textbf{Irrelevant}: Medically irrelevant information covered in the consultation, e.g. the pet of a patient with a suspected chest infection just died.
\end{itemize}

\noindent The definitions of critical, non-critical and irrelevant information are taken from previous work on human evaluation of generated clinical notes \citep{moramarco-etal-2022-human, savkov-etal-2022-consultation}.

In short, notes generated by ICL are strongly preferred over notes generated by the fine-tuned model and, on average, \textit{slightly} preferred over the human-written notes (\autoref{tab:human-eval}), validating the high performance reported by the automatic metrics. We note, however, that inter-annotator agreement is low and speculate why this might be in \textsection \ref{limitations}.

\section{Related Work}

Automated note generation from doctor-patient conversations has received increasing attention in recent years \citep{finley-etal-2018-automated, enarvi-etal-2020-generating, Molenaar2020MedicalDS, knoll-etal-2022-user}. Different methods have been proposed, such as extractive-abstractive approaches \citep{joshi-etal-2020-dr, krishna-etal-2021-generating, Su2022ExtractAA} and fine-tuning PLMs (\citealt{zhang-etal-2021-leveraging-pretrained}, similar to our approach in \textsection \ref{fine-tuning}). Others have focused on curating data for training and benchmarking \citep{papadopoulos-korfiatis-etal-2022-primock57}, including the use of LLMs to produce synthetic data \citep{Chintagunta2021MedicallyAG}. Lastly, there have been efforts to improve the evaluation of generated clinical notes, both with automatic metrics \citep{moramarco-etal-2022-human} and human evaluation \citep{savkov-etal-2022-consultation}. While recent literature has commented on the \textit{potential} of ICL for note generation \citep{Lee2023BenefitsLA}, our work is among the first to evaluate this approach rigorously.

\section{Conclusion}

We present our submission to the MEDIQA-Chat shared task for clinical note generation from doctor-patient dialogues. We evaluated a fine-tuning-based approach with LED and an ICL-based approach with GPT-4, ranking second and first, respectively, among all submissions. Human evaluation with three physicians revealed that notes produced by GPT-4 via ICL were strongly preferred over notes produced by LED and, on average, slightly preferred over human-written notes. We conclude that ICL is a promising path toward clinical note generation from doctor-patient conversations.

\section*{Limitations}
\label{limitations}

\paragraph{Evaluation of generated text is difficult} Evaluating automatically generated text, including clinical notes, is generally hard due to the inherently subjective nature of many aspects of output quality. Automatic evaluation metrics such as ROUGE and BERTScore are imperfect \citep{deutsch-etal-2022-examining} and may not correlate with aspects of expert judgment. However, they are frequently used to evaluate model-generated clinical notes and do correlate with certain aspects of quality \citep{moramarco-etal-2022-human}. To further validate our findings, we also conducted a human evaluation with three expert physicians (\textsection \ref{results:human-eval}). As noted previously \citep{savkov-etal-2022-consultation}, even human evaluation of clinical notes is far from perfect; inter-annotator agreement is generally low, likely because physicians have differing opinions on the importance of each patient statement and whether it should be included in a consultation note. We also found low inter-annotator agreement in our human evaluation and speculate this is partially due to differences in specialties among the physicians. Physicians 1 and 3, both from family medicine, had high agreement with each other but low agreement with physician 2 (cardiac surgery, see \autoref{tab:human-eval}). Investigating better automatic metrics and best practices for evaluating clinical notes (and generated text more broadly) is an active field of research. We hope to integrate novel and performant metrics in future work.

\paragraph{Data privacy}

While our GPT-4 based solution achieves the best performance, it is not compliant with data protection regulations such as HIPAA; although Azure does advertise a HIPAA-compliant option.\footnote{\url{https://azure.microsoft.com/en-us/products/cognitive-services/openai-service\#security}} From a privacy perspective, locally deploying a model such as LED may be preferred; however, our results suggest that more work is needed for this approach to reach acceptable performance (see \autoref{tab:human-eval}). In either case, when implementing automated clinical note-generation systems, healthcare providers and developers should ensure that the whole system --- including text-to-speech, data transmission \& storage, and model inference --- adheres to privacy and security requirements to maintain trust and prevent privacy violations in the clinical setting.

\section*{Ethics Statement}

Developing an automated system for clinical note generation from doctor-patient conversations raises several ethical considerations. First, informed consent is crucial: patients must be made aware of their recording, and data ownership must be prioritized. Equitable access is also important; the system must be usable for patients from diverse backgrounds, including those with disabilities, limited technical literacy, or language barriers. Addressing issues of data bias and fairness are necessary to avoid unfair treatment or misdiagnosis for certain patient groups. The system must implement robust security measures to protect patient data from unauthorized access or breaches. Establishing clear lines of accountability for errors or harms arising from using an automated system for note generation is paramount. Disclosure of known limitations or potential risks associated with using the system is essential to maintain trust in the patient-physician relationship. Finally, ongoing evaluations are necessary to ensure that system performance does not degrade and negatively impact the quality of care.

\section*{Acknowledgements}

This research was enabled in part by support provided by the Digital Research Alliance of Canada (\href{https://alliancecan.ca/}{alliancecan.ca}) and Compute Ontario (\href{https://www.computeontario.ca/}{www.computeontario.ca}). We thank all internal and external reviewers for their thoughtful feedback, which improved earlier drafts of this manuscript.

\section*{Author Contributions}
\label{sec:contrib}

John Giorgi (JG), Augustin Toma (AT) and Ronald Xie (RX) led the project in general, including data cleaning and processing, model implementation, and running experiments. JG wrote the initial manuscript and designed the human evaluation with feedback from AT and RX. AT and RX recruited Sondra~S.~Chen, Kevin~R.~An and Grace~X.~Zheng, who served as expert physicians in the human evaluation. Bo Wang provided high-level feedback and advice.

\bibliography{anthology,custom}
\bibliographystyle{acl_natbib}

\clearpage

\appendix

\section{Subtask A}
\label{appendix:subtask-a}

In subtask A of the Dialogue2Note Summarization shared task, given a partial doctor-patient dialogue, the goals are to: (1) predict the appropriate section header, e.g. ``PASTMEDICALHX'' and (2) generate that specific section of a note. We approached this task by fine-tuning a PLM on the provided training set, following a canonical, sequence-to-sequence training process (see \autoref{seq2seq} for details). In preliminary experiments, we found that the instruction-tuned FLAN-T5 \citep{Chung2022ScalingIL} performed particularly well at this task.

\begin{figure}[t]
\includegraphics[width=\columnwidth]{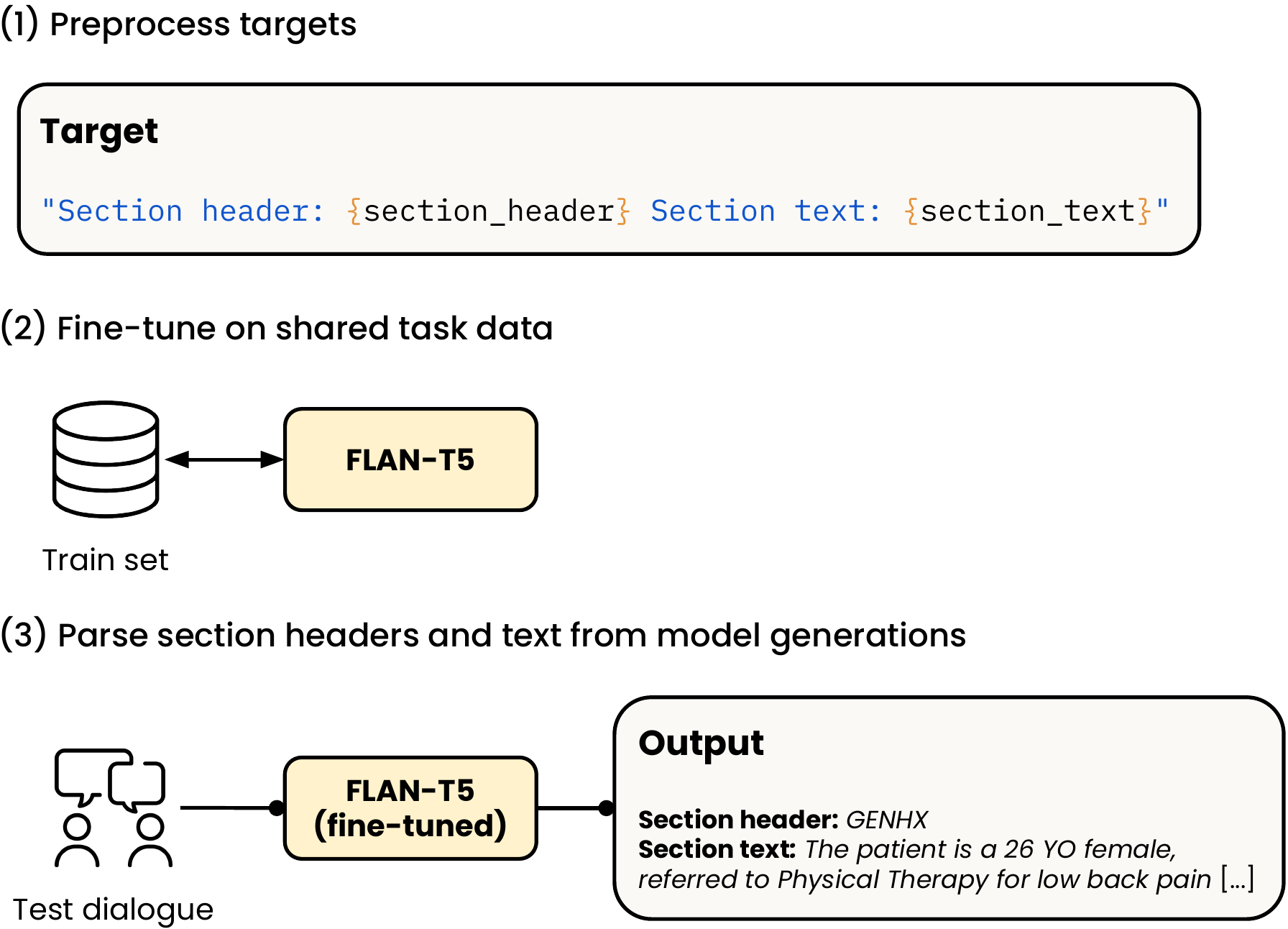}
\caption{Fine-tuning FLAN-T5 \citep{Chung2022ScalingIL} for subtask A. Before training, targets are preprocessed as ``Section header: \texttt{\{section\_header\}} Section text: \texttt{\{section\_text\}}''. After decoding, the section header and text are parsed using regular expressions.}
\label{fig:flan-t5}
\end{figure}

We hypothesized that jointly learning to predict the section header and generate the section text would improve overall performance. To do this, we preprocessed the training set so the targets were of the form: ``Section header: \texttt{\{section\_header\}} Section text: \texttt{\{section\_text\}}''. After decoding, the section header and text were parsed using regular expressions and evaluated separately (\autoref{fig:flan-t5}). Section header prediction was evaluated as the fraction of predicted headers that match the ground truth (accuracy), and section text was evaluated similarly to subtask B (see \textsection \ref{evaluation}). In cases where the model output an invalid section header,\footnote{In practice, we found that the fine-tuned model rarely, if ever, generates invalid section headers} we replaced it with ``GENHX'' (general history), which tends to summarize the contents of the other sections. The model was fine-tuned on a single NVIDIA A100-40GB GPU. Hyperparameters were lightly tuned on the validation set (\autoref{tab:taska-hyperparameters}).

\begin{figure}[t]
\centering
\includegraphics[width=\columnwidth]{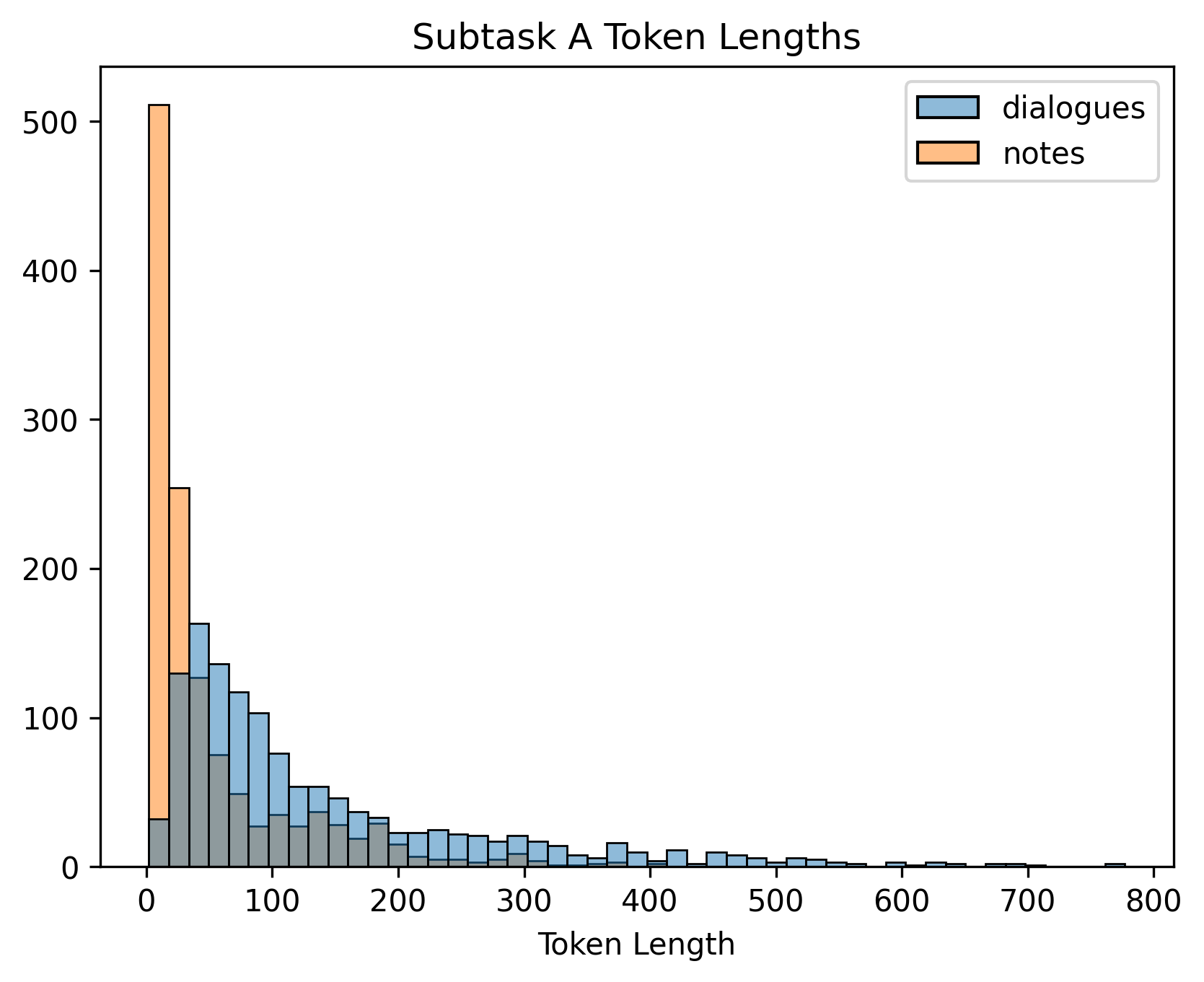}
\caption{Histogram of token lengths for subtask A train and validation sets. Dialogues and notes were tokenized with \href{https://github.com/huggingface/tokenizers}{HuggingFace Tokenizers} using ``\texttt{google/flan-t5-large}''. Lengths greater than the 99th-percentile are omitted to make the plot legible.}
\label{fig:taska-lengths}
\end{figure}

We present the results of our approach on the validation set in \autoref{tab:flan-results}. Similar to subtask B (see \textsection \ref{results:fine-tuning}), we find, perhaps unsurprisingly, that scaling the model size from FLAN-T5\textsubscript{BASE} (24 layers, \(\sim\)250M parameters) to FLAN-T5\textsubscript{LARGE} (48 layers, \(\sim\)780M parameters) leads to large improvements in performance. Performance is further improved by jointly learning to predict section headers and generate note sections. Our submission to the shared task based on this approach tied for first on section header prediction (78\% accuracy), and ranked first for note section generation (average ROUGE-1, BERTScore and BLEURT F1-score of 57.9).

\begin{table*}[t]
	\small
	\centering
	\caption{Hyperparameters used with FLAN-T5 on the Dialogue2Note subtask A}
	\label{tab:taska-hyperparameters}
	\small
	\begin{tabular}{p{0.19\textwidth} p{0.31\textwidth} p{0.46\textwidth}}
		\toprule
		Hyperparameter                    & Value                                   & Comment                                                                                                                             \\ \midrule
		\texttt{max\_source\_length}      & 1024                                    & truncate input sequences to this max length                                                                                         \\
		\texttt{max\_target\_length}      & 512                                    & truncate output sequences to this max length                                                                                        \\
		\texttt{source\_prefix} &
		\textit{``Summarize the following patient-doctor dialogue. Include all medically relevant information, including family history, diagnosis, past medical (and surgical) history, immunizations, lab results and known allergies. You should first predict the most relevant clinical note section header and then summarize the dialogue. Dialogue:''} & instruction text prepended to all inputs \\
		\texttt{train\_batch\_size}       & 8                                       & batch size during training                                                                                                          \\
		\texttt{eval\_batch\_size}        & 12                                       & batch size during inference                                                                                                         \\
		\texttt{learning\_rate}           & 1e-4                                    & learning rate during training                                                                                                       \\
		\texttt{optimizer}                & AdamW \citep{Loshchilov2017DecoupledWD} & optimizer used during training                                                                                                      \\
		\texttt{num\_train\_epochs}       & 20                                      & total number of training epochs                                                                                                     \\
		\texttt{warmup\_ratio}            & 0.1                                     & proportion of training steps to linearly increase the learning rate to \texttt{learning\_rate}                                      \\
		\texttt{lr\_scheduler}            & linear with warmup                      & learning rate linearly increased during first \texttt{warmup\_ratio} fraction of train steps and linearly decreased to 0 afterwords \\
		\texttt{weight\_decay}            & 0.01                                    & not applied to bias \& \texttt{LayerNorm} weights                                                                                   \\
		\texttt{label\_smoothing} & 0.1                                     & label smoothing factor used during training                                                                                         \\
		\texttt{bf16}                     & true                                    & whether to use BF16 during training                                                                                                 \\
		\texttt{num\_beams}               & 2                                       & beam size used during beam search decoding                                                                                          \\  \bottomrule
	\end{tabular}
\end{table*}

\begin{table*}[t]
\centering
\small
\caption{Fine-tuning FLAN-T5. Accuracy of predicted section headers and score of generated note sections is shown. Jointly learning to predict section headers and generate notes improve performance. \textbf{Bold}: best scores.}
\label{tab:flan-results}
\begin{tabular}{@{}lccccc@{}}
\toprule
                            &                           & \multicolumn{4}{c}{Note generation} \\ \cmidrule(l){3-6} 
Model                       & Header prediction (\%) & ROUGE-1 F1   & BERTScore F1   & BLEURT   & Avg.   \\ \midrule
Random header               & 8.0                       & --        & --          & --       & --     \\
Majority header             & 22.0                      & --        & --          & --       & --     \\
FLAN-T5\textsubscript{BASE} & 71.0                      & 40.1      & 70.5        & 52.7     & 54.5   \\
FLAN-T5\textsubscript{LARGE}                                & \textbf{79.0} & \textbf{49.8} & \textbf{74.5} & \textbf{58.0} & \textbf{60.8} \\
\quad \(\hookrightarrow\) \textit{w/o header prediction} & --            & 48.0          & 74.3          & 57.6          & 59.9          \\ \bottomrule
\end{tabular}
\end{table*}

\begin{table*}[t]
	\small
	\centering
	\caption{Hyperparameters used with Longformer-Encoder-Decoder (LED) on the Dialogue2Note subtask B}
	\label{tab:led-hyperparameters}
	\small
	\begin{tabular}{p{0.19\textwidth} p{0.31\textwidth} p{0.46\textwidth}}
		\toprule
		Hyperparameter                    & Value                                   & Comment                                                                                                                             \\ \midrule
		\texttt{max\_source\_length}      & 4096                                    & truncate input sequences to this max length                                                                                         \\
		\texttt{max\_target\_length}      & 1024                                    & truncate output sequences to this max length                                                                                        \\
		\texttt{source\_prefix} &
		\textit{``Summarize the following patient-doctor dialogue. Include all medically relevant information, including family history, diagnosis, past medical (and surgical) history, immunizations, lab results and known allergies. Dialogue:''} & instruction text prepended to all inputs \\
		\texttt{train\_batch\_size}       & 8                                       & batch size during training                                                                                                          \\
		\texttt{eval\_batch\_size}        & 6                                       & batch size during inference                                                                                                         \\
		\texttt{learning\_rate}           & 3e-5                                    & learning rate during training                                                                                                       \\
		\texttt{optimizer}                & AdamW \citep{Loshchilov2017DecoupledWD} & optimizer used during training                                                                                                      \\
		\texttt{num\_train\_epochs}       & 50                                      & total number of training epochs                                                                                                     \\
		\texttt{warmup\_ratio}            & 0.1                                     & proportion of training steps to linearly increase the learning rate to \texttt{learning\_rate}                                      \\
		\texttt{lr\_scheduler}            & linear with warmup                      & learning rate linearly increased during first \texttt{warmup\_ratio} fraction of train steps and linearly decreased to 0 afterwords \\
		\texttt{weight\_decay}            & 0.01                                    & not applied to bias \& \texttt{LayerNorm} weights                                                                                   \\
		\texttt{label\_smoothing} & 0.1                                     & label smoothing factor used during training                                                                                         \\
		\texttt{fp16}                     & true                                    & whether to use FP16 during training                                                                                                 \\
		\texttt{num\_beams}               & 4                                       & beam size used during beam search decoding                                                                                          \\ 
		\texttt{min\_length}              & 100                                     & min length of generated sequences                                                                                                   \\
		\texttt{max\_length}              & 1024                                    & max length of generated sequences                                                                                                   \\
		\texttt{length\_penalty}          & 2.0                                     & values \(> 0\) promote longer output sequences                                                                                      \\ 
		\texttt{no\_repeat\_ngram}  & 3                                       & ngrams of this size can only occur once                                                                                             \\ \bottomrule
	\end{tabular}
\end{table*}

\section{Subtask B}
\label{appendix:subtask-b}

\subsection{Hyperparameter tuning of LED}
\label{appendix:led-hyperparams}

We lightly tuned the hyperparameters of LED\textsubscript{LARGE-PubMed} on the subtask B validation set against the average ROUGE-1 F1, BERTScore F1 and BLEURT-20 scores. The best hyperparameters obtained are given in \autoref{tab:led-hyperparameters}. We used the same hyperparameters when fine-tuning LED\textsubscript{BASE} and LED\textsubscript{LARGE} in \textsection \ref{results:fine-tuning}.

\subsection{Post processing LEDs outputs}
\label{appendix:led-post-process}

In practice, we found that the fine-tuned LED model sometimes produces invalid section headers; notably, this problem did not occur with the ICL-based approach using GPT-4. Therefore, we lightly post-processed LEDs outputs using a simple script that identifies section headers produced by the model not in the ground truth set and uses fuzzy string matching\footnote{We used \url{https://github.com/seatgeek/thefuzz}} to replace them with the closest valid header. For example, in one run, this process converted the (incorrect) predicted section header ``HISTORY OF PRESENT'' to the nearest valid header ``HISTORY OF PRESENT ILLNESS''.

\section{Fine-tuning Seq2Seq Models}
\label{seq2seq}

When training the sequence-to-sequence (seq2seq) models for both subtask A (\autoref{appendix:subtask-a}) and B (\textsection \ref{fine-tuning}), we followed a canonical supervised fine-tuning (SFT) process. We start with a pre-trained, encoder-decoder transformer-based language model \citep{NIPS2017_3f5ee243}. First, the encoder maps each token in the input to a contextual embedding. Then, the autoregressive decoder generates an output, token-by-token, attending to the outputs of the encoder at each timestep. Decoding proceeds until a special ``end-of-sequence'' token (e.g. \texttt{</s>}) is generated, or a maximum number of tokens have been generated. Formally, \(X\) is the \textit{input} sequence, which in our case is a doctor-patient dialogue, and \(Y\) is the corresponding \textit{output} sequence of length \(T\), in our case a clinical note. We model the conditional probability:

\begin{align}
    p(Y | X) = \prod_{t=1}^Tp(y_t | X, y_{<t})    
\end{align}

\noindent During training, we optimize over the model parameters \(\theta\) the sequence cross-entropy loss:

\begin{align}
    \ell(\theta) = - \sum_{t=1}^T \log p(y_t | X, y_{<t} ; \theta) \label{cross-entropy-loss}
\end{align}

\noindent maximizing the log-likelihood of the training data. As is common, we use \textit{teacher forcing} during training, feeding previous ground truth inputs to the decoder when predicting the next token in the sequence. During inference, we generate the output using beam search \citep{graves2012sequence}. Beams are ranked by mean token log probability after applying a length penalty. Models are fine-tuned using the HuggingFace Transformers library.\footnote{\url{https://github.com/huggingface/transformers/blob/main/examples/pytorch/summarization/run_summarization.py}}

\end{document}